%% file: IGARSS2025LaTeXTemplate.tex
\documentclass[conference,a4paper]{IEEEtran}
\IEEEoverridecommandlockouts

\usepackage[hidelinks]{hyperref}
\usepackage[cmex10]{amsmath}
\usepackage{amssymb,amsfonts}
\usepackage{dblfloatfix}

\usepackage[ruled,vlined]{algorithm2e}
\usepackage{graphicx}
\graphicspath{{Figures/PDF/}{Figures/PNG/}}

\usepackage[numbers,compress]{natbib}
\usepackage{texnames}
\usepackage{bm,bbm}
\usepackage{orcidlink}
\usepackage{subcaption}
\usepackage{xcolor}
\usepackage{array}
\usepackage{multirow}
\newcolumntype{C}[1]{>{\centering\let\newline\\\arraybackslash\hspace{0pt}}m{#1}}
\usepackage{float}
\usepackage{hyperref}

\begin{document}
\title{\LARGE{TRANSFORMING VOLCANIC MONITORING: A DATASET AND BENCHMARK FOR ONBOARD VOLCANO ACTIVITY DETECTION}
\vspace{-2mm}
\thanks{This work has been supported by the SmartSat CRC, whose activities are funded by the Australian Government’s CRC Program.}
}
\author{\IEEEauthorblockN{\textit{Darshana Priyasad, Tharindu Fernando, Maryam Haghighat, Harshala Gammulle, Clinton Fookes}}\vspace{2mm}
    \IEEEauthorblockA{Signal Processing, Artificial Intelligence and Vision Technologies, Queensland University of Technology, Brisbane, Australia}
}
\maketitle

\input{TEX_Abstract}

\input{TEX_Introduction}
\input{TEX_Dataset}

\input{TEX_Results}
\input{TEX_Conclusion}

\small
\bibliographystyle{IEEEtranN}

\end{document}

%% file: TEX_Abstract.tex
\begin{abstract}
Natural disasters, such as volcanic eruptions, pose significant challenges to daily life and incur considerable global economic losses. The emergence of next-generation small-satellites, capable of constellation-based operations, offers unparalleled opportunities for near-real-time monitoring and onboard processing of such events. However, a major bottleneck remains the lack of extensive annotated datasets capturing volcanic activity, which hinders the development of robust detection systems. This paper introduces a novel dataset explicitly designed for volcanic activity and eruption detection, encompassing diverse volcanoes worldwide. The dataset provides binary annotations to identify volcanic anomalies or non-anomalies, covering phenomena such as temperature anomalies, eruptions, and volcanic ash emissions. These annotations offer a foundational resource for developing and evaluating detection models, addressing a critical gap in volcanic monitoring research. Additionally, we present comprehensive benchmarks using state-of-the-art models to establish baselines for future studies. Furthermore, we explore the potential for deploying these models onboard next-generation satellites. Using the Intel Movidius Myriad X VPU as a testbed, we demonstrate the feasibility of volcanic activity detection directly onboard. This capability significantly reduces latency and enhances response times, paving the way for advanced early warning systems. This paves the way for innovative solutions in volcanic disaster management, encouraging further exploration and refinement of onboard monitoring technologies.
\end{abstract}


%% file: TEX_Introduction.tex
\section{Introduction}
\label{sec:introduction}

The commissioning of Kanyini, Australia's first home-grown satellite, and other AI-enabled satellites such as $\phi$sat2 \cite{marin2021phi} have opened up numerous opportunities for onboard processing of remotely sensed imagery, facilitating a wide range of applications \cite{lu2024onboard, di2022early, del2021board, razzano2024ai, vatsal2024continuous, guerrisi2022convolutional}. Among these, natural disaster monitoring has gained significant attention, particularly in light of the increasing frequency of such events due to climate change \cite{casagli2023landslide, xing2023flood}. While volcanic eruptions are not directly linked to climate change, they still represent a major natural disaster with substantial financial implications \cite{mota2024monitoring, girina2023monitoring}, and near real-time monitoring of these events can significantly reduce financial losses by enabling more effective disaster risk management strategies.

Recent advancements in deep learning (DL) have enabled the extraction of robust features from multi-spectral and hyper-spectral remotely sensed images, leading to improved performance \cite{wang2021review, thangavel2023autonomous, yuan2021deep}. However, training deep networks for generalized inference requires large, high-quality datasets \cite{alzubaidi2023survey}. Existing publicly available datasets for volcanic activity detection are relatively small \cite{del2021board}, underscoring the need for larger datasets with more comprehensive annotations of volcanic eruptions and anomalies.

To address this gap, we present a new dataset annotated for volcanic activity detection, derived from publicly available Sentinel-2 L1C product data \cite{CDSE}. To assess the potential of DL methods for volcanic activity detection, we benchmarked our dataset using widely adopted computer vision models. In line with the specific requirements for onboard deployment, including uplink bandwidth limitations and minimum performance thresholds, we employed score-level knowledge distillation. This approach enabled the transfer of knowledge from larger, more complex teacher models to a simpler, more efficient DCNN-based student network that met all deployment criteria. To further ensure the feasibility of onboard deployment, we tested the optimized student model on the Intel Movidius Myriad X Vision Processing Unit (VPU), the processing hardware used in the Kanyini satellite. The data collection and annotation process is detailed in Section \ref{sec:data_collection} while benchmarking results are discussed in Section \ref{sec:benchmarking}.

%% file: TEX_Dataset.tex
\section{Dataset Collection : Methodology}
\label{sec:data_collection}

Onboard volcanic activity detection offers a promising solution for global monitoring, addressing limitations in near-real-time processing and downlink capacities of satellite data. However, training and robust evaluation require large-scale, ready-to-use datasets. Despite the abundance of satellite products, challenges persist due to the limited availability of free, high-resolution (low-GSD) data and the effort-intensive process of downloading, preprocessing, and manual annotation. 
This section provides a detailed overview of the entire workflow, starting from the process of downloading satellite imagery to the steps involved in preprocessing and annotating the data for volcanic activity detection.

\subsection{Volcano Selection and Data Retrieval}
\label{subsec:volcano_selection}

The selection of volcanoes for the dataset was guided by two primary criteria. First, the NCEI/WDS Global Significant Volcanic Eruptions Database was utilized to identify volcanoes with recent eruptions, specifically those with a Volcanic Explosivity Index (VEI) of 2 or higher (only terrestrial volcanic activity). Second, the Smithsonian Institution's Global Volcanism Program database \cite{Smithsonian} was used to include volcanoes exhibiting diverse activity types, such as fumarolic emissions, even in the absence of recent eruptions. Following these criteria, 35 volcanoes were selected including the dates of activity, representing diverse geographic locations across six inhabited continents, as illustrated in Figure \ref{fig:volcano_distribution}.

We selected Sentinel-2 Level-1C products as the primary data source due to their free availability (since 2016) and comparatively higher spatial resolution. Sentinel-2 offers a best-case Ground Sampling Distance (GSD) of 10 m (most bands at 2 0m), whereas Landsat primarily provides 30 m GSD, with only a single panchromatic band at 15m. Level-1C products, representing Top-of-Atmosphere (TOA) reflectance, were chosen over Level-2A (Bottom-of-Atmosphere reflectance) because atmospheric correction cannot be performed onboard a satellite due to technical and computational constraints. This guarantees that the proposed models align with the data captured during onboard inference.

To identify relevant data, we first determined the Sentinel-2 tiles corresponding to each volcano using the Sentinel-2 UTM tiling grid provided by the eAtlas catalogue. A Python script was employed to automate the download of scenes from the Copernicus Data Space Ecosystem (CDSE) based on tile IDs. Data was filtered to include a maximum of 50 scenes acquired within 10 days before and up to 1 year after an eruption. Cloud cover was not used as a filtering criterion, as a volcano might still be visible despite partial cloud obstruction. For highly active volcanoes, such as Etna and Stromboli, the search window was extended to ensure comprehensive coverage.

\subsection{Preprocessing and Annotation}
\label{subsec:preprocess_annotation}

\begin{figure}[!t]
\centering
\includegraphics[width=0.98\linewidth]{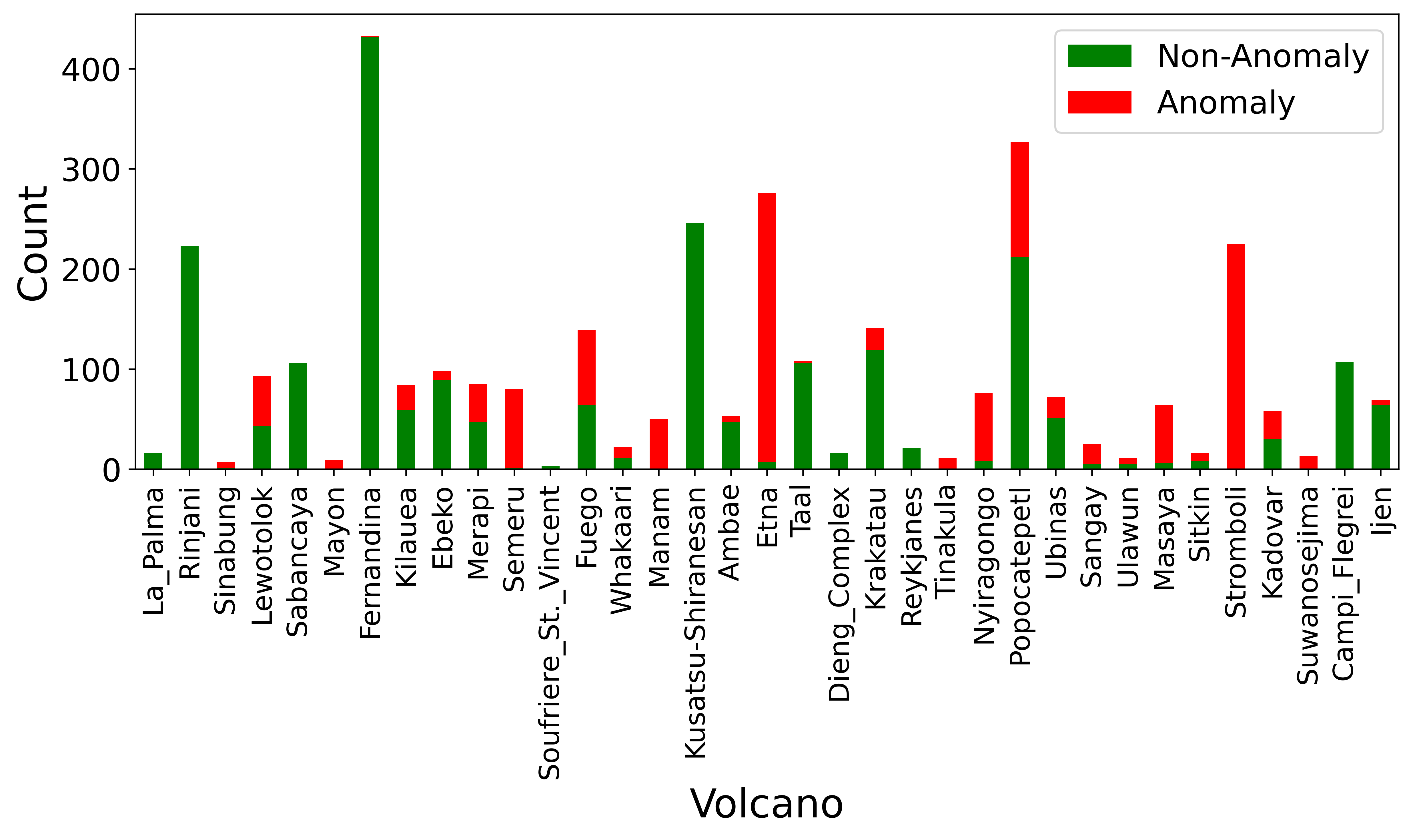} 
\caption{Distribution of anomaly and non-anomaly samples across the selected volcanoes in the dataset (2016-2024). }
\label{fig:volcano_distribution}
\vspace{-5mm}
\end{figure}

Each downloaded Sentinel-2 scene consists of separate JP2000 raster images for 13 spectral bands (visible, NIR, and SWIR) provided as Digital Numbers (DN). While DN values typically range between 0 and 10,000, saturation effects in certain pixels can exceed this range. To address this, bands were normalized by dividing by 10,000 and then clipped to a maximum value of 1 to manage saturated pixels effectively. For SWIR-augmented RGB data, the clipping was performed after augmentation to preserve essential spectral information.

To create the dataset, a sample RGB image for each tile ID corresponding to a volcano was selected under low-cloud conditions, combining bands B04, B03, and B02 (R, G, B). Volcano center coordinates were determined manually using GIMP. In cases where multiple volcanoes were visible within a tile, all were included as part of the same volcano. 

\begin{figure*}[!ht]
    \centering
    \begin{subfigure}[b]{0.161\textwidth}
        \centering
        \includegraphics[width=\textwidth]{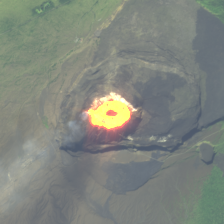}
    \end{subfigure}
    \hfill
    \begin{subfigure}[b]{0.161\textwidth}
        \centering
        \includegraphics[width=\textwidth]{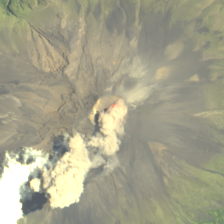}
    \end{subfigure}
    \hfill
    \begin{subfigure}[b]{0.161\textwidth}
        \centering
        \includegraphics[width=\textwidth]{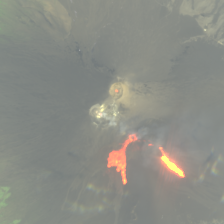}
    \end{subfigure}
    \hfill
    \begin{subfigure}[b]{0.161\textwidth}
        \centering
        \includegraphics[width=\textwidth]{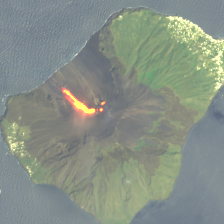}
    \end{subfigure}
    \hfill
    \begin{subfigure}[b]{0.161\textwidth}
        \centering
        \includegraphics[width=\textwidth]{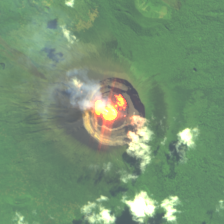}
    \end{subfigure}
    \hfill
    \begin{subfigure}[b]{0.161\textwidth}
        \centering
        \includegraphics[width=\textwidth]{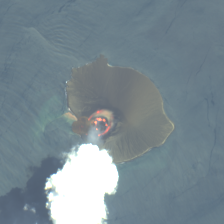}
    \end{subfigure}
    \vfill\vspace{1mm}
    \begin{subfigure}[b]{0.161\textwidth}
        \centering
        \includegraphics[width=\textwidth]{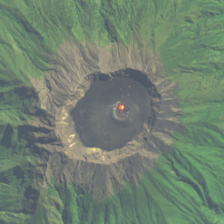}
    \end{subfigure}
    \hfill
    \begin{subfigure}[b]{0.161\textwidth}
        \centering
        \includegraphics[width=\textwidth]{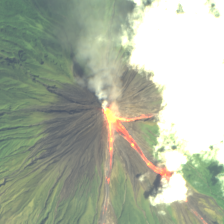}
    \end{subfigure}
    \hfill
    \begin{subfigure}[b]{0.161\textwidth}
        \centering
        \includegraphics[width=\textwidth]{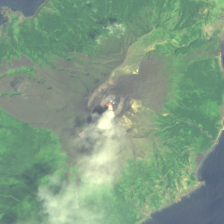}
    \end{subfigure}
    \hfill
    \begin{subfigure}[b]{0.161\textwidth}
        \centering
        \includegraphics[width=\textwidth]{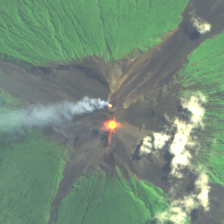}
    \end{subfigure}
    \hfill
    \begin{subfigure}[b]{0.161\textwidth}
        \centering
        \includegraphics[width=\textwidth]{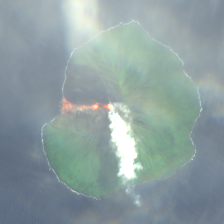}
    \end{subfigure}
    \hfill
    \begin{subfigure}[b]{0.161\textwidth}
        \centering
        \includegraphics[width=\textwidth]{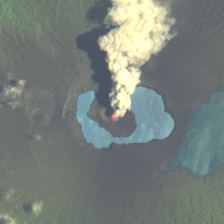}
    \end{subfigure}
    \caption{Sample SWIR-augmented RGB images (high-temperature regions are highlighted in red \cite{del2021board}) of volcanic activity labelled as anomalies. Images with visible fumes and tephra have also been categorised as anomalies.}
    \label{fig:sample_volcano}
    \vspace{-5mm}
\end{figure*}

Using these coordinates, 224×224 pixel SWIR-augmented RGB images (3-channel) were generated as in Equation \ref{eq:swir_aug_rgb}, centering the volcano, as described in \cite{del2021board}. These images were manually inspected to remove occluded volcanoes and were annotated for anomalies and non-anomalies. Images, where volcanic fumes were indistinguishable from surrounding clouds, were excluded to maintain dataset consistency and quality.

\begin{equation}
\begin{aligned}
RED = 2.5 \times B04 + max(0, B12 - 0.1) \\
GREEN = 2.5 \times B03 + max(0, B11 - 0.1) \\
BLUE = 2.5 \times B02
\end{aligned}
\label{eq:swir_aug_rgb}
\end{equation}

The dataset includes both SWIR-augmented RGB images and 9-channel MSI cubes, created by concatenating bands B02–B12 (excluding B08, B09, and B10). This approach addresses the limitations of SWIR augmentation, which may overlook critical information in other bands, such as NIR. By incorporating both SWIR-augmented RGB images and full 9-band MSI cubes, the dataset offers a comprehensive resource for volcanic activity detection and further benchmarking. Additionally, the dataset features SWIR-augmented RGB images and MSI cubes at three different Ground Sampling Distances (GSD): 10 m, 20 m, and 75 m (75 m is similar to Kanyini's GSD). The original 20 m GSD Sentinel-2 bands were interpolated to create 10 m and 75 m GSD versions, then center-cropped to position the volcano at the image center. Each resulting image has a resolution of 224×224 pixels, with the volcano centered.


The resulting dataset comprises 3,383 samples, including 2,153 non-anomaly (NA) and 1,230 anomaly (A) images, sourced from 35 volcanoes, each with SWIR-augmented SWIR and 9-channel MSI cubles at 10 m, 20 m, and 75 m GSD resolutions. The class distribution across these volcanoes is shown in Figure \ref{fig:volcano_distribution}. The dataset exhibits class imbalance due to the exclusion of images with volcanic occlusions, which are common, and the rarity of anomalies and eruptions. For holdout evaluations, the dataset was randomly split into training, validation, and test sets with 2,343 (1,493 NA, 850 A), 311 (201 NA, 110 A), and 729 (459 NA, 270 A) samples, respectively. This represents a 2.5-fold increase in sample size compared to the state-of-the-art dataset \cite{del2021board}, which consisted of 1,317 images (910 NA, 407 A). Selected SWIR-augmented RGB image samples are shown in Figure \ref{fig:sample_volcano}.


%% file: TEX_Results.tex
\section{Benchmarking : Results and Discussion}
\label{sec:benchmarking}

In this section, we present the benchmark results for the proposed dataset, as shown in Table \ref{tab:benchmark_performance}. Given the class imbalance in the dataset and the critical importance of accurately detecting anomalous volcanic activity compared to non-active states, we selected the F1 score and Precision-Recall Area Under the Curve (PR-AUC) as the primary performance metrics, which provide a more balanced evaluation of model performance, particularly in scenarios where the detection of anomalies is more critical.

\subsection{Model Selection for Benchmarking}
\label{subsec:model_selection}

We approached detecting volcanic anomalies as a binary classification problem within the domain of computer vision. To this end, we selected state-of-the-art (SOTA) models, including ResNet variants \cite{he2016deep} and transformer-based architectures \cite{dosovitskiy2020image, liu2021swin}, for benchmarking. However, considering the constraints on uplink bandwidth and onboard processing capabilities in small-satellites, we also incorporated smaller, more efficient models like MobileNet \cite{howard2019searching} for comparison. We used transfer learning for all models, using pre-trained weights on ImageNet \cite{deng2009imagenet}. The first convolutional layer of each model was adjusted to accommodate the 9-channel input data (MSI cube), for relevant evaluations.

\begin{table*}[ht]
\renewcommand{\arraystretch}{1.10}
\centering
\caption{Benchmark performance of volcano anomaly detection on the proposed dataset. The best three performances for each setting are highlighted in \textcolor{red}{red}, \textcolor{blue}{blue} and \textcolor{teal}{teal}, respectively. Only the top-10 benchmark SOTA models are included in the table. DCNN is a custom lightweight network designed to meet onboard constraints;the distilled model uses knowledge distillation.}

\label{tab:benchmark_performance}
\begin{tabular}{|C{2.5cm}|C{0.81cm}|C{0.81cm}|C{0.81cm}|C{0.81cm}|C{0.81cm}|C{0.81cm}||C{0.81cm}|C{0.81cm}|C{0.81cm}|C{0.81cm}|C{0.81cm}|C{0.81cm}|} 
\hline
\multirow{3}{*}{Model} & \multicolumn{6}{c||}{Holdout Validation} & \multicolumn{6}{c|}{LOVO Validation}\\\cline{2-13}
& \multicolumn{2}{c|}{10 m GSD} & \multicolumn{2}{c|}{20 m GSD} & \multicolumn{2}{c||}{75 m GSD} & \multicolumn{2}{c|}{10 m GSD} & \multicolumn{2}{c|}{20 m GSD} & \multicolumn{2}{c|}{75 m GSD} \\\cline{2-13}
& F1 & AUC & F1 & AUC & F1 & AUC & F1 & AUC & F1 & AUC & F1 & AUC \\\hline\hline
\multicolumn{13}{|c|}{SWIR Augmented RGB} \\\hline

MobileNetV3(L) \cite{howard2019searching} & 0.9416 & 0.9822 & 0.9573 & 0.9837 & 0.8832 & \textcolor{teal}{0.9671} & \textcolor{blue}{0.9721} & \textcolor{blue}{0.9877} & 0.9454 & 0.9751 & 0.7933 & 0.9104 \\\hline

MobileNetV3(S) \cite{howard2019searching} & 0.9466 & 0.9839 & \textcolor{teal}{0.9575} & 0.9886 & \textcolor{teal}{0.9124} & \textcolor{blue}{0.9710} & 0.9620 & 0.9823 & 0.9470 & 0.9784 & 0.7419 & 0.8225\\\hline

Resnet18 \cite{he2016deep} & 0.9450 & 0.9868 & \textcolor{blue}{0.9585} & \textcolor{blue}{0.9901} & \textcolor{blue}{0.9175} & \textcolor{red}{0.9765} & 0.9617 & 0.9845 & 0.9478 & 0.9695 & \textcolor{teal}{0.8641} & \textcolor{blue}{0.9218}\\\hline

Resnet50 \cite{he2016deep} & 0.9364 & 0.9826 & 0.9346 & 0.9769 & 0.9043 & 0.9567 & 0.9639 & \textcolor{teal}{0.9857} & 0.9327 & 0.9694 & \textcolor{blue}{0.8643} & \textcolor{teal}{0.9167}\\\hline

ResNet101 \cite{he2016deep} & 0.9486 & 0.9833 & 0.9357 & 0.9806 & \textcolor{red}{0.9225} & 0.9589 & 0.9455 & 0.9710 & \textcolor{teal}{0.9522} & \textcolor{teal}{0.9805} & 0.8100 & 0.8687\\\hline

ResNet152 \cite{he2016deep} & 0.9446 & \textcolor{teal}{0.9892} & 0.9436 & 0.9841 & 0.8772 & 0.9507 & 0.9560 & 0.9780 & 0.9395 & 0.9606 & \textcolor{red}{0.8673} & \textcolor{red}{0.9265}\\\hline

ViT-Base \cite{dosovitskiy2020image} & \textcolor{teal}{0.9506} & 0.9843 & 0.9268 & 0.9802 & 0.8906 & 0.9432 & 0.9323 & 0.9689 & 0.8362 & 0.8868 & 0.5526 & 0.5971 \\\hline


Swin-Tiny \cite{liu2021swin} & 0.9470 & 0.9861 & 0.9474 & \textcolor{teal}{0.9888} & 0.8979 & 0.9602 & \textcolor{teal}{0.9693} & 0.9831 & 0.9296 & 0.9624 & 0.6919 & 0.7516 \\\hline

Swin-Base \cite{liu2021swin} & \textcolor{red}{0.9564} & \textcolor{blue}{0.9907} & 0.9497 & \textcolor{red}{0.9913} & 0.8743 & 0.9408 & 0.9613 & 0.9803 & \textcolor{red}{0.9609} & \textcolor{red}{0.9834} & 0.7288 & 0.7902\\\hline

Swin-Large \cite{liu2021swin} & \textcolor{blue}{0.9551} & \textcolor{red}{0.9909} & \textcolor{red}{0.9586} & 0.9872 & 0.9060 & 0.9631 & \textcolor{red}{0.9724} & \textcolor{red}{0.9894} & \textcolor{blue}{0.9542} & \textcolor{blue}{0.9832} & 0.7408 & 0.7753 \\\hline\hline

DCNN & 0.9213 & 0.9799 & 0.9211 & 0.9777 & 0.8861 & 0.9555 & 0.9451 & 0.9565 & 0.9274 & 0.9275 & 0.6422 & 0.6791 \\\hline

DCNN-Distilled & 0.9328 & 0.9859 & 0.9185 & 0.9818 & 0.9098 & 0.9723 & 0.9443 & 0.9597 & 0.9304 & 0.9422 & 0.8135 & 0.8603 \\\hline\hline

\multicolumn{13}{|c|}{9-Band Multispectral Cube} \\\hline

MobileNetV3(L) & \textcolor{red}{0.9645} & \textcolor{red}{0.9916} & \textcolor{teal}{0.9517} & 0.9842 & 0.8958 & 0.9655 & 0.9688 & 0.9808 & 0.9540 & \textcolor{blue}{0.9840} & 0.7976 & 0.8437\\\hline

MobileNetV3(S) & 0.9468 & \textcolor{teal}{0.9873} & 0.9362 & \textcolor{blue}{0.9899} & 0.9025 & 0.9565 & 0.9655 & 0.9762 & 0.9571 & 0.9777 & \textcolor{teal}{0.8303} & 0.8670\\\hline

Resnet18 & \textcolor{blue}{0.9545} & \textcolor{blue}{0.9889} & \textcolor{red}{0.9631} & \textcolor{red}{0.9911} & \textcolor{red}{0.9410} & \textcolor{blue}{0.9780} & 0.9722 & \textcolor{blue}{0.9849} & \textcolor{teal}{0.9690} & \textcolor{teal}{0.9814} & \textcolor{blue}{0.8424} & 0.8495\\\hline

Resnet50 & 0.9464 & 0.9837 & 0.9502 & 0.9853 & 0.9124 & 0.9681 & \textcolor{red}{0.9774} & \textcolor{red}{0.9893} & 0.9596 & 0.9739 & 0.8067 & \textcolor{blue}{0.8781}\\\hline

ResNet101 & 0.9482 & 0.9841 & \textcolor{blue}{0.9524} & 0.9834 & 0.9214 & 0.9571 & \textcolor{teal}{0.9752} & 0.9826 & 0.9651 & 0.9756 & 0.8020 & \textcolor{teal}{0.8671}\\\hline

ResNet152 & 0.9439 & 0.9807 & 0.9284 & 0.9823 & 0.9145 & 0.9568 & 0.9718 & \textcolor{teal}{0.9846} & 0.9587 & \textcolor{red}{0.9843} & \textcolor{red}{0.8586} & \textcolor{red}{0.9013} \\\hline

ViT-Base & 0.9231 & 0.9758 & 0.9456 & 0.9849 & 0.8852 & 0.9463 & 0.9295 & 0.9557 & 0.8593 & 0.8789 & 0.7880 & 0.7918\\\hline


Swin-Tiny & 0.9493 & 0.9822 & 0.9443 & \textcolor{teal}{0.9863} & \textcolor{blue}{0.9314} & \textcolor{teal}{0.9779} & \textcolor{teal}{0.9752} & 0.9811 & \textcolor{blue}{0.9696} & 0.9733 & 0.7428 & 0.7884\\\hline

Swin-Base & 0.9373 & 0.9656 & 0.9382 & 0.9754 & \textcolor{teal}{0.9280} & 0.9763 & \textcolor{blue}{0.9757} & 0.9787 & \textcolor{red}{0.9706} & 0.9779 & 0.7541 & 0.8184\\\hline

Swin-Large & \textcolor{teal}{0.9504} & 0.9810 & 0.9377 & 0.9780 & 0.9159 & \textcolor{red}{0.9788} & \textcolor{red}{0.9774} & 0.9797 & 0.9679 & 0.9776 & 0.6788 & 0.7797 \\\hline\hline

DCNN & 0.9213 & 0.9679 & 0.9174 & 0.9626 & 0.8952 & 0.9657 & 0.9437 & 0.9434 & 0.9641 & 0.9660 & 0.7554 & 0.8161 \\\hline

DCNN-Distilled & 0.9373 & 0.9888 & 0.9515 & 0.9814 & 0.9055 & 0.9717 & 0.9456 & 0.9558 & 0.9715 & 0.9846 & 0.6912 & 0.8560 \\\hline

\end{tabular}
\vspace{-5mm}
\end{table*}

\subsection{Experimental Setup}
\label{subsec:experiment_setup}

We utilized both holdout validation (as described in \ref{subsec:preprocess_annotation}) and Leave-One-Volcano-Out (LOVO) cross-validation to ensure robust and comprehensive evaluations. The volcanic activity detection task was assessed using both 3-channel SWIR-augmented RGB images and 9-channel MSI data, evaluated at resolutions of 10 m, 20 m, and 75 m. For LOVO evaluations, since some volcanoes contained samples from only a single class (as illustrated in Figure \ref{fig:volcano_distribution}), performance metrics such as PR-AUC and F1 score were computed by aggregating predictions across all volcanoes, rather than averaging metrics from individual volcano results. ResNet and MobileNet models were trained with the Adam optimizer \cite{kingma2014adam}, while ViT and Swin Transformer models employed the SGD optimizer, using a learning rate of 0.001. The training was conducted for 20 epochs using the BCEWithLogitsLoss.

\vspace{-1mm}
\subsection{Results Analysis and Discussion}
\label{subsec:results_discussion}

In the holdout validation results, Swin-Large and Swin-Base models demonstrated superior performance with SWIR-augmented RGB images, while ResNet18 achieved the best results with 9-band MSI data. Across resolutions, both 10 m and 20 m GSD data consistently outperformed 75 m GSD inputs. This performance disparity is attributed to the smaller appearance of volcanoes in 75 m GSD imagery, where anomalous activity is confined to a few pixels, limiting feature learning. Conversely, higher-resolution data (10 m and 20 m) provides greater pixel coverage of volcanic craters and surroundings, enabling more distinct anomaly detection. When comparing SWIR-augmented RGB data and 9-band MSI inputs, both 10 m and 20 m GSD datasets achieved comparable best performances. However, at 75 m GSD, the 9-band MSI data showed a substantial performance improvement. This is likely due to the inclusion of complementary spectral features across bands, which enhances the network’s ability to learn discriminative features despite the coarser resolution. As expected, LOVO validation demonstrated lower performance compared to holdout validation across all scenarios, due to the more challenging nature of cross-volcano evaluations. Swin Transformer models consistently outperformed others with both input data types, underscoring the strengths of transformer-based architectures in this domain. Additionally, PR-AUC values were significantly high in all settings, despite the imbalanced class distribution (low anomaly representation). This indicates the models' effectiveness in detecting the minority class (anomalies), a critical capability for applications like volcanic eruption detection, where rare events need reliable identification. These findings affirm that deep learning methods are well-suited for anomaly detection in volcanic activity.

\subsection{Knowledge Distillation from ResNet18 to DCNN}
\label{subsec:distillation}

Prior work on onboard satellite image processing, such as CloudScout \cite{giuffrida2020cloudscout} deployed on the HyperScout-2 satellite \cite{esposito2019orbit}, emphasized key requirements for deep networks in space applications. These include a memory footprint of no more than 5 MB to accommodate uplink bandwidth constraints and a minimum accuracy threshold of 85\%. While most benchmarked models met the accuracy criterion, their sizes were impractical for onboard deployment.

To address these constraints, we designed a lightweight deep convolutional neural network (DCNN) with a 4.92 MB memory footprint. The architecture included an initial Conv2D block (16 filters) followed by batch normalization, max pooling, and three subsequent convolution blocks (32, 64, and 128 filters), each comprising three Conv2D layers, batch normalization, and max pooling. The flattened output was passed through a multi-layer perceptron (MLP) with decreasing units (32, 16, 8) and a final classification layer. Training configurations were consistent with other benchmark models for comparability. 

While the DCNN satisfied size requirements, its performance lagged across all configurations due to limited depth, particularly under Leave-One-Volcano-Out (LOVO) validation, where it failed to meet the 85\% accuracy threshold (see DCNN in Table \ref{tab:benchmark_performance}). To improve performance, we employed knowledge distillation \cite{allen2020towards, gou2021knowledge}, transferring knowledge from the ResNet18 model (teacher network) (overall best benchmark model as in Table \ref{tab:benchmark_performance}) to the DCNN (student network) using score-level distillation. The distillation process utilized a combined loss of RMSE and BCEWithLogitsLoss. The distilled-DCNN achieved a notable performance improvement, as depicted in Table \ref{tab:benchmark_performance} (DCNN-Distilled), meeting both size and accuracy criteria (accuracy of 0.8773) for SWIR-augmented RGB images, making it suitable for onboard deployment.

\vspace{-2mm}
\subsection{Evaluations on Intel Movidius Myriad X VPU}
\label{subsec:onboard}

The Kanyini satellite uses an Intel Movidius Myriad X VPU for onboard processing, differing from the on-ground hardware typically used for training models. To ensure consistency across different hardware, PyTorch models were converted to ONNX format and executed on the VPU (using Intel Neural Compute Stick 2) using OpenVINO, with performance metrics compared. Under holdout validation, the 10 m, 20 m, and 75 m GSD input configurations achieved F1 scores of 0.9305$\textcolor{red}{\downarrow}$, 0.9222$\textcolor{teal}{\uparrow}$, and 0.8949$\textcolor{red}{\downarrow}$ with SWIR-augmented RGB images (18 ms inference time), and 0.9267$\textcolor{red}{\downarrow}$, 0.9350$\textcolor{red}{\downarrow}$, and 0.9077$\textcolor{teal}{\uparrow}$ with a 9-band MSI cube (25 ms inference time). The slight performance variations in VPU evaluations ($<1\%$ average variation) were attributed to differences in floating-point operations on different hardware. 

%% file: TEX_Conclusion.tex
\section{Conclusion}
\label{sec:colclusion}
This study demonstrates the potential for real-time volcanic activity detection onboard next-generation satellites, addressing the challenge of limited datasets with a novel Sentinel-2 L1C-based dataset covering 35 volcanoes across six continents. Benchmarking results validate the effectiveness of deep learning models under various configurations, while knowledge distillation enables smaller, efficient networks suitable for deployment. Testing on the Intel Movidius Myriad X VPU confirms the feasibility of onboard implementation, paving the way for robust and scalable satellite-based volcanic monitoring solutions. In future, this research can be extended to include volcanic-event forecasting and real-time monitoring within small satellite constellations, enabling continuous observation and rapid response, thereby improving impact mitigation.

%% file: IGARSS2025LaTeXTemplate.bbl
\begin{thebibliography}{27}
\providecommand{\natexlab}[1]{#1}
\providecommand{\url}[1]{#1}
\csname url@samestyle\endcsname
\providecommand{\newblock}{\relax}
\providecommand{\bibinfo}[2]{#2}
\providecommand{\BIBentrySTDinterwordspacing}{\spaceskip=0pt\relax}
\providecommand{\BIBentryALTinterwordstretchfactor}{4}
\providecommand{\BIBentryALTinterwordspacing}{\spaceskip=\fontdimen2\font plus
\BIBentryALTinterwordstretchfactor\fontdimen3\font minus \fontdimen4\font\relax}
\providecommand{\BIBforeignlanguage}[2]{{%
\expandafter\ifx\csname l@#1\endcsname\relax
\typeout{** WARNING: IEEEtranN.bst: No hyphenation pattern has been}%
\typeout{** loaded for the language `#1'. Using the pattern for}%
\typeout{** the default language instead.}%
\else
\language=\csname l@#1\endcsname
\fi
#2}}
\providecommand{\BIBdecl}{\relax}
\BIBdecl

\bibitem[Marin et~al.(2021)Marin, Coelho, Deconinck, Babkina, Longepe, and Pastena]{marin2021phi}
A.~Marin, C.~Coelho, F.~Deconinck, I.~Babkina, N.~Longepe, and M.~Pastena, ``Phi-sat-2: Onboard ai apps for earth observation,'' \emph{Proc. Space Artif. Intell}, 2021.

\bibitem[Lu et~al.(2024)Lu, Jones, Zhao, Sun, Qin, Liu, Li, Abeysekara, Mueller, Oliver, et~al.]{lu2024onboard}
S.~Lu, E.~Jones, L.~Zhao, Y.~Sun, K.~Qin, J.~Liu, J.~Li, P.~Abeysekara, N.~Mueller, S.~Oliver \emph{et~al.}, ``Onboard ai for fire smoke detection using hyperspectral imagery: an emulation for the upcoming kanyini hyperscout-2 mission,'' \emph{IEEE Journal of Selected Topics in Applied Earth Observations and Remote Sensing}, 2024.

\bibitem[Di~Stasio et~al.(2022)Di~Stasio, Sebastianelli, Meoni, and Ullo]{di2022early}
P.~Di~Stasio, A.~Sebastianelli, G.~Meoni, and S.~L. Ullo, ``Early detection of volcanic eruption through artificial intelligence on board,'' in \emph{2022 IEEE International Conference on Metrology for Extended Reality, Artificial Intelligence and Neural Engineering (MetroXRAINE)}.\hskip 1em plus 0.5em minus 0.4em\relax IEEE, 2022, pp. 714--718.

\bibitem[Del~Rosso et~al.(2021)Del~Rosso, Sebastianelli, Spiller, Mathieu, and Ullo]{del2021board}
M.~P. Del~Rosso, A.~Sebastianelli, D.~Spiller, P.~P. Mathieu, and S.~L. Ullo, ``On-board volcanic eruption detection through cnns and satellite multispectral imagery,'' \emph{Remote Sensing}, vol.~13, no.~17, p. 3479, 2021.

\bibitem[Razzano et~al.(2024)Razzano, Di~Stasio, Mauro, Meoni, Esposito, Schirinzi, and Ullo]{razzano2024ai}
F.~Razzano, P.~Di~Stasio, F.~Mauro, G.~Meoni, M.~Esposito, G.~Schirinzi, and S.~L. Ullo, ``Ai techniques for near real-time monitoring of contaminants in coastal waters on board future phisat-2 mission,'' \emph{arXiv preprint arXiv:2404.19586}, 2024.

\bibitem[Vatsal et~al.(2024)Vatsal, Nandi, and Manilal]{vatsal2024continuous}
V.~Vatsal, G.~Nandi, and P.~Manilal, ``Continuous monitoring for road flooding with satellite onboard computing for navigation for orbitalai $\phi$sat-2 challenge,'' \emph{arXiv preprint arXiv:2405.02868}, 2024.

\bibitem[Guerrisi et~al.(2022)Guerrisi, Del~Frate, and Schiavon]{guerrisi2022convolutional}
G.~Guerrisi, F.~Del~Frate, and G.~Schiavon, ``Convolutional autoencoder algorithm for on-board image compression,'' in \emph{IGARSS 2022-2022 IEEE International Geoscience and Remote Sensing Symposium}.\hskip 1em plus 0.5em minus 0.4em\relax IEEE, 2022, pp. 151--154.

\bibitem[Casagli et~al.(2023)Casagli, Intrieri, Tofani, Gigli, and Raspini]{casagli2023landslide}
N.~Casagli, E.~Intrieri, V.~Tofani, G.~Gigli, and F.~Raspini, ``Landslide detection, monitoring and prediction with remote-sensing techniques,'' \emph{Nature Reviews Earth \& Environment}, vol.~4, no.~1, pp. 51--64, 2023.

\bibitem[Xing et~al.(2023)Xing, Yang, Zan, Dong, Yao, Liu, and Zhang]{xing2023flood}
Z.~Xing, S.~Yang, X.~Zan, X.~Dong, Y.~Yao, Z.~Liu, and X.~Zhang, ``Flood vulnerability assessment of urban buildings based on integrating high-resolution remote sensing and street view images,'' \emph{Sustainable Cities and Society}, vol.~92, p. 104467, 2023.

\bibitem[Mota et~al.(2024)Mota, Pacheco, Pimentel, and Gil]{mota2024monitoring}
R.~Mota, J.~M. Pacheco, A.~Pimentel, and A.~Gil, ``Monitoring volcanic plumes and clouds using remote sensing: A systematic review,'' \emph{Remote Sensing}, vol.~16, no.~10, p. 1789, 2024.

\bibitem[Girina et~al.(2023)Girina, Manevich, Loupian, Uvarov, Korolev, Sorokin, Romanova, Kramareva, and Burtsev]{girina2023monitoring}
O.~Girina, A.~Manevich, E.~Loupian, I.~Uvarov, S.~Korolev, A.~Sorokin, I.~Romanova, L.~Kramareva, and M.~Burtsev, ``Monitoring the thermal activity of kamchatkan volcanoes during 2015--2022 using remote sensing,'' \emph{Remote Sensing}, vol.~15, no.~19, p. 4775, 2023.

\bibitem[Wang et~al.(2021)Wang, Liu, Liu, Zhu, Hou, Liu, and Li]{wang2021review}
C.~Wang, B.~Liu, L.~Liu, Y.~Zhu, J.~Hou, P.~Liu, and X.~Li, ``A review of deep learning used in the hyperspectral image analysis for agriculture,'' \emph{Artificial Intelligence Review}, vol.~54, no.~7, pp. 5205--5253, 2021.

\bibitem[Thangavel et~al.(2023)Thangavel, Spiller, Sabatini, Amici, Sasidharan, Fayek, and Marzocca]{thangavel2023autonomous}
K.~Thangavel, D.~Spiller, R.~Sabatini, S.~Amici, S.~T. Sasidharan, H.~Fayek, and P.~Marzocca, ``Autonomous satellite wildfire detection using hyperspectral imagery and neural networks: A case study on australian wildfire,'' \emph{Remote Sensing}, vol.~15, no.~3, p. 720, 2023.

\bibitem[Yuan et~al.(2021)Yuan, Zhuang, Schaefer, Feng, Guan, and Fang]{yuan2021deep}
K.~Yuan, X.~Zhuang, G.~Schaefer, J.~Feng, L.~Guan, and H.~Fang, ``Deep-learning-based multispectral satellite image segmentation for water body detection,'' \emph{IEEE Journal of Selected Topics in Applied Earth Observations and Remote Sensing}, vol.~14, pp. 7422--7434, 2021.

\bibitem[Alzubaidi et~al.(2023)Alzubaidi, Bai, Al-Sabaawi, Santamar{\'\i}a, Albahri, Al-dabbagh, Fadhel, Manoufali, Zhang, Al-Timemy, et~al.]{alzubaidi2023survey}
L.~Alzubaidi, J.~Bai, A.~Al-Sabaawi, J.~Santamar{\'\i}a, A.~S. Albahri, B.~S.~N. Al-dabbagh, M.~A. Fadhel, M.~Manoufali, J.~Zhang, A.~H. Al-Timemy \emph{et~al.}, ``A survey on deep learning tools dealing with data scarcity: definitions, challenges, solutions, tips, and applications,'' \emph{Journal of Big Data}, vol.~10, no.~1, p.~46, 2023.

\bibitem[Agency()]{CDSE}
\BIBentryALTinterwordspacing
E.~S. Agency, ``Sentinel-2.'' [Online]. Available: \url{https://dataspace.copernicus.eu/explore-data/data-collections/sentinel-data/sentinel-2}
\BIBentrySTDinterwordspacing

\bibitem[Global Volcanism~Program(2024)]{Smithsonian}
\BIBentryALTinterwordspacing
.~Global Volcanism~Program, ``[database] volcanoes of the world,'' 2024, distributed by Smithsonian Institution, compiled by Venzke, E. [Online]. Available: \url{https://doi.org/10.5479/si.GVP.VOTW5-2024.5.2}
\BIBentrySTDinterwordspacing

\bibitem[He et~al.(2016)He, Zhang, Ren, and Sun]{he2016deep}
K.~He, X.~Zhang, S.~Ren, and J.~Sun, ``Deep residual learning for image recognition,'' in \emph{Proceedings of the IEEE conference on computer vision and pattern recognition}, 2016, pp. 770--778.

\bibitem[Dosovitskiy(2020)]{dosovitskiy2020image}
A.~Dosovitskiy, ``An image is worth 16x16 words: Transformers for image recognition at scale,'' \emph{arXiv preprint arXiv:2010.11929}, 2020.

\bibitem[Liu et~al.(2021)Liu, Lin, Cao, Hu, Wei, Zhang, Lin, and Guo]{liu2021swin}
Z.~Liu, Y.~Lin, Y.~Cao, H.~Hu, Y.~Wei, Z.~Zhang, S.~Lin, and B.~Guo, ``Swin transformer: Hierarchical vision transformer using shifted windows,'' in \emph{Proceedings of the IEEE/CVF international conference on computer vision}, 2021, pp. 10\,012--10\,022.

\bibitem[Howard et~al.(2019)Howard, Sandler, Chu, Chen, Chen, Tan, Wang, Zhu, Pang, Vasudevan, et~al.]{howard2019searching}
A.~Howard, M.~Sandler, G.~Chu, L.-C. Chen, B.~Chen, M.~Tan, W.~Wang, Y.~Zhu, R.~Pang, V.~Vasudevan \emph{et~al.}, ``Searching for mobilenetv3,'' in \emph{Proceedings of the IEEE/CVF international conference on computer vision}, 2019, pp. 1314--1324.

\bibitem[Deng et~al.(2009)Deng, Dong, Socher, Li, Li, and Fei-Fei]{deng2009imagenet}
J.~Deng, W.~Dong, R.~Socher, L.-J. Li, K.~Li, and L.~Fei-Fei, ``Imagenet: A large-scale hierarchical image database,'' in \emph{2009 IEEE conference on computer vision and pattern recognition}.\hskip 1em plus 0.5em minus 0.4em\relax Ieee, 2009, pp. 248--255.

\bibitem[Kingma(2014)]{kingma2014adam}
D.~P. Kingma, ``Adam: A method for stochastic optimization,'' \emph{arXiv preprint arXiv:1412.6980}, 2014.

\bibitem[Giuffrida et~al.(2020)Giuffrida, Diana, de~Gioia, Benelli, Meoni, Donati, and Fanucci]{giuffrida2020cloudscout}
G.~Giuffrida, L.~Diana, F.~de~Gioia, G.~Benelli, G.~Meoni, M.~Donati, and L.~Fanucci, ``Cloudscout: A deep neural network for on-board cloud detection on hyperspectral images,'' \emph{Remote Sensing}, vol.~12, no.~14, p. 2205, 2020.

\bibitem[Esposito et~al.(2019)Esposito, Conticello, Pastena, and Dom{\'\i}nguez]{esposito2019orbit}
M.~Esposito, S.~S. Conticello, M.~Pastena, and B.~C. Dom{\'\i}nguez, ``In-orbit demonstration of artificial intelligence applied to hyperspectral and thermal sensing from space,'' in \emph{CubeSats and SmallSats for remote sensing III}, vol. 11131.\hskip 1em plus 0.5em minus 0.4em\relax SPIE, 2019, pp. 88--96.

\bibitem[Allen-Zhu and Li(2020)]{allen2020towards}
Z.~Allen-Zhu and Y.~Li, ``Towards understanding ensemble, knowledge distillation and self-distillation in deep learning,'' \emph{arXiv preprint arXiv:2012.09816}, 2020.

\bibitem[Gou et~al.(2021)Gou, Yu, Maybank, and Tao]{gou2021knowledge}
J.~Gou, B.~Yu, S.~J. Maybank, and D.~Tao, ``Knowledge distillation: A survey,'' \emph{International Journal of Computer Vision}, vol. 129, no.~6, pp. 1789--1819, 2021.

\end{thebibliography}
